\documentclass[pdflatex,sn-nature,oneside]{sn-jnl}% Style for submissions to Nature Portfolio journals
\usepackage{graphicx}%
\usepackage{multirow}%
\usepackage{amsmath,amssymb,amsfonts}%
\usepackage{amsthm}%
\usepackage[title]{appendix}%
\usepackage{xcolor}%
\usepackage{textcomp}%
\usepackage{manyfoot}%
\usepackage{booktabs}%
\usepackage{algorithm}%
\usepackage{algorithmicx}%
\usepackage{algpseudocode}%
\usepackage{listings}%
\makeatletter
\@mparswitchfalse
\makeatother

\theoremstyle{thmstyleone}%
\theoremstyle{thmstyletwo}%

\theoremstyle{thmstylethree}%

\begin{document}

\title[Intersectional Fairness in VLMs]{Intersectional Fairness in Vision-Language Models for Medical Image Disease Classification}


\author[1]{\fnm{Yupeng} \sur{Zhang}}\email{yzha3836@uni.sydney.edu.au}

\author[2]{\fnm{Adam G.} \sur{Dunn}}\email{adam.dunn@sydney.edu.au}

\author*[3]{\fnm{Usman} \sur{Naseem}}\email{usman.naseem@mq.edu.au}

\author*[1]{\fnm{Jinman} \sur{Kim}}\email{jinman.kim@sydney.edu.au}

\affil*[1]{\orgdiv{Biomedical Data Analysis and Visualisation (BDAV) Lab, School of Computer Science}, \orgname{The University of Sydney}, \orgaddress{\city{Sydney}, \state{NSW}, \country{Australia}}}

\affil[2]{\orgdiv{Sydney School of Public Health}, \orgname{The University of Sydney}, \orgaddress{\city{Sydney}, \state{NSW}, \country{Australia}}}

\affil[3]{\orgdiv{School of Computing}, \orgname{Macquarie University}, \orgaddress{\city{Sydney}, \state{NSW}, \country{Australia}}}


\abstract{Medical artificial intelligence (AI) systems, particularly multimodal vision-language models (VLM), often exhibit intersectional biases where models are systematically less confident in diagnosing marginalised patient subgroups. Such bias can lead to higher rates of inaccurate and missed diagnoses due to demographically skewed data and divergent distributions of diagnostic certainty. Current fairness interventions frequently fail to address these gaps or compromise overall diagnostic performance to achieve statistical parity among the subgroups. In this study, we developed Cross-Modal Alignment Consistency (CMAC-MMD), a training framework that standardises diagnostic certainty across intersectional patient subgroups. Unlike traditional debiasing methods, this approach equalises the model's decision confidence without requiring sensitive demographic data during clinical inference. We evaluated this approach using 10,015 skin lesion images (HAM10000) with external validation on 12,000 images (BCN20000), and 10,000 fundus images for glaucoma detection (Harvard-FairVLMed), stratifying performance by intersectional age, gender, and race attributes. In the dermatology cohort, the proposed method reduced the overall intersectional missed diagnosis gap (difference in True Positive Rate, $\Delta$TPR) from 0.50 to 0.26 while improving the overall Area Under the Curve (AUC) from 0.94 to 0.97 compared to standard training. Similarly, for glaucoma screening, the method reduced $\Delta$TPR from 0.41 to 0.31, achieving a better AUC of 0.72 (vs. 0.71 baseline). This establishes a scalable framework for developing high-stakes clinical decision support systems that are both accurate and can perform equitably across diverse patient subgroups, ensuring reliable performance without increasing privacy risks.}




\maketitle

\section{Introduction}\label{sec1}

Melanoma and glaucoma represent typical cases in which early detection directly determines patient survival and the prevention of irreversible vision loss. In melanoma, substantial racial disparities in outcomes have been documented; notably, reported five-year survival rates range from 66–70\% for Black patients compared to 90–94\% for White patients, a disparity that has widened and persisted in the modern treatment era \cite{Hu2006, Dawes2016}. This gap is driven by diagnostic delay, since 52\% of Black patients are diagnosed at an advanced stage compared to 16\% of White patients \cite{Hu2006}. In glaucoma, the leading cause of irreversible blindness among Black Americans, prevalence is more than double that of White populations \cite{Tielsch1991, Varma2004, Sommer1991}, and Black individuals are six to eight times more likely to experience blindness from the disease \cite{Varma2004, Allison2021}. Hispanic and Latino populations face undiagnosed rates ranging from 62\% to 75\% \cite{Quigley2001, Varma2004}, representing missed opportunities for intervention during the decade-long window before symptomatic progression.

Artificial intelligence (AI) assisted screening in primary care and community settings offers the most scalable pathway to address these access barriers. US Food and Drug Administration (FDA)-approved autonomous AI systems have demonstrated the feasibility of point-of-care detection without specialist review \cite{Abramoff2018, Venkatesh2024}, and community deployments have identified previously undiagnosed diseases in up to 26\% of screened populations \cite{EyeArt2024}. However, current AI systems trained predominantly on light-skinned populations exhibit substantial performance degradation on darker skin tones \cite{DDI, groh2021evaluating}. Critically, when AI assistance was provided to primary care physicians for skin lesion diagnosis, the accuracy gap between light and dark skin increased by five percentage points \cite{Groh2024}, demonstrating that AI designed without accounting for population bias can exacerbate rather than reduce healthcare disparities. Vision-language models (VLM), which integrate medical imaging with clinical text and represent the current state-of-the-art for multimodal diagnostic support \cite{CLIP, BLIP2, BioMedCLIP, PMC-CLIP, MedCLIP, PubMedCLIP, Zhang2024BiomedGPT}, inherit these risks. Driven by their reliance on component encoders pre-trained on demographically skewed or uncurated datasets, these architectures frequently embed latent biases that compromise downstream performance. Consequently, recent evidence indicates that VLMs systematically underdiagnose marginalised subgroups, including intersectional populations such as Black female patients, at rates exceeding those of human radiologists \cite{Yang2025}.

%% Figure 1: Intersectional bias illustration
\begin{figure}[ht]
\centering
\includegraphics[width=\textwidth]{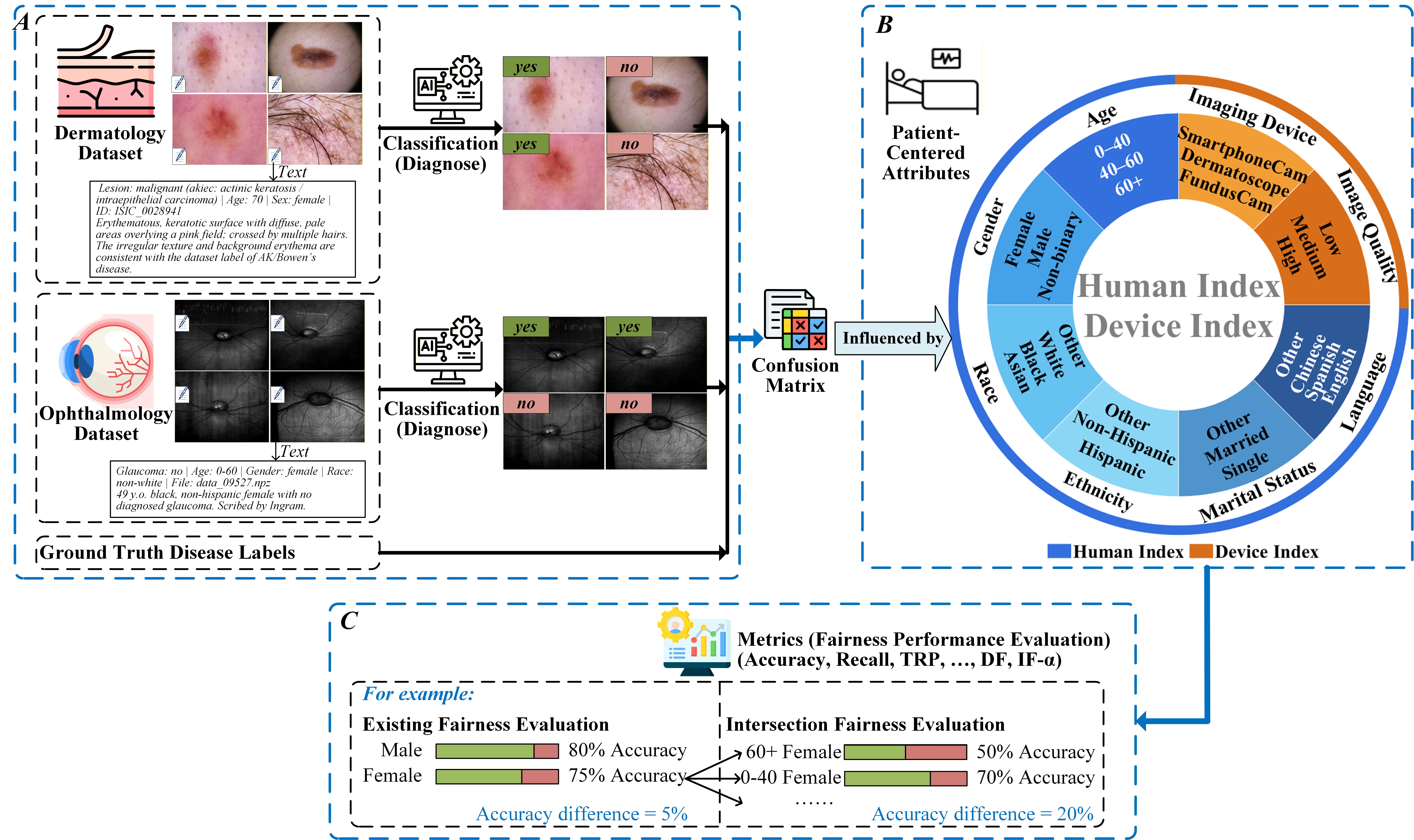}
\caption{\textbf{Intersectional fairness unmasks hidden performance disparities in multimodal medical AI.} \textbf{A} Schematic of the VLM workflow for dermatology (top) and ophthalmology (bottom), where images and clinical text are processed to generate diagnostic predictions, which are aggregated into a confusion matrix. \textbf{B} The ``Intersectional Wheel'' defines the compounding variables that influence diagnostic performance. These factors are stratified into the Human Index (patient-centred attributes) and the Device Index (technical factors). \textbf{C} A comparative evaluation example derived from the confusion matrix. A model may appear fair when performance is averaged across single-attribute analyses, but this can mask hidden algorithmic bias in clinical decision support.}\label{fig1}
\end{figure}

A fundamental limitation of current fairness research is its focus on single demographic attributes, including race, gender, or age, which are evaluated in isolation \cite{Liu2025fairness, Usman2023}. However, patients exist at demographic intersections where biases compound (Fig.~\ref{fig1}). A model that appears fair when evaluated by gender alone may exhibit substantially larger disparities for specific intersectional subgroups such as elderly Black women. Moreover, existing fairness interventions often produce a ``levelling down'' effect, achieving statistical parity by degrading performance for all groups rather than improving outcomes for disadvantaged populations \cite{McCradden2020, Chen2021}, an ethically untenable trade-off in clinical practice. We identify a critical, often overlooked mechanism underlying these failures is the ``certainty gap'': even when models achieve similar aggregate accuracy, they often exhibit systematic disparities in diagnostic confidence for underrepresented groups, leaving these patients in a ``grey zone'' of uncertainty that makes diagnoses unstable and vulnerable to missed detection. This certainty gap is not captured by conventional fairness metrics that evaluate only the final classification outcomes.

In this study, we introduce Cross-Modal Alignment Consistency via Maximum Mean Discrepancy (CMAC-MMD) to promote equitable medical image classification across intersectional patient subgroups. The fundamental premise of our approach is that equitable diagnostic AI must produce equally confident predictions for all patients, regardless of their demographic profile. This consistency is achieved by directly regularising the distribution of diagnostic certainty scores across intersectional subgroups during training on CLIP-based architectures, rather than attempting to debias high-dimensional feature representations. Our examination of CMAC-MMD's efficacy spanned two clinical domains: skin lesion classification using dermatology cohorts (HAM10000 and BCN20000) and glaucoma detection using an ophthalmology cohort (Harvard-FairVLMed). We aimed to reduce intersectional disparities in missed diagnoses while maintaining overall diagnostic performance. Importantly, demographic attributes are not required during inference, thereby preserving patient privacy. We compared CMAC-MMD against common fairness methods that incorporated several strategies: resampling to balance data representation across groups; reweighting to adjust sample importance; and adversarial training techniques to learn subgroup-invariant representations. We used the area under the receiver operating characteristic curve (AUC) and the difference in true positive rate ($\Delta$TPR) to analyse overall screening performance and disparities in missed diagnoses. Furthermore, we employed Differential Fairness (DF) and Intersectional Fairness-alpha (IF-$\alpha$) \cite{Foulds2020, Gaurav2023} to quantify fairness across intersectional subgroups and validated generalisability using an external dataset.

\section{Methods}\label{sec2}

\subsection{Study Design and Clinical Cohort Selection}\label{subsec1}

This study was designed as a retrospective multi-cohort evaluation to assess whether the proposed CMAC-MMD method reduces intersectional diagnostic disparities while maintaining overall classification performance across two distinct clinical domains: dermatology (skin lesion classification) and ophthalmology (glaucoma detection). We selected datasets that met two strict inclusion criteria for intersectional fairness analysis: (1) availability of at least two demographic attributes to construct intersectional subgroups, and (2) sufficient sample size within each resulting subgroup (minimum 100 samples) to ensure statistically reliable metric estimation \cite{RicciLara2022}.

\textbf{Dermatology Cohorts} The primary dermatology dataset was HAM10000 \cite{DVN/DBW86T_2018}, comprising 10,015 dermoscopic images of pigmented skin lesions with associated age and gender metadata. We constructed six intersectional subgroups by stratifying age into three clinically informed bins (0-40, 41-60, and 60+ years) crossed with binary gender. This age stratification reflects established risk inflection points in dermatology: the 0-40 bin represents a baseline population, while the 41-60 and 60+ bins capture cohorts where melanoma risk accelerates substantially, consistent with evidence of major biomolecular shifts in skin metabolism around age 44 \cite{StanfordMedicine2024aging} and the use of age 60 as a primary prognostic threshold in the American Joint Committee on Cancer (AJCC) Melanoma Staging Database \cite{DVN/DBW86T_2018}. The dataset was split into training ($n = 5{,}968$; 60\%), validation ($n = 1{,}989$; 20\%), and held-out test ($n = 1{,}989$; 20\%) sets using stratified sampling to preserve subgroup proportions. External validation was performed on BCN20000 \cite{BCN}, an independent dataset of approximately 12,000 labeled dermoscopic images, to assess generalisability under distribution shift.

\textbf{Ophthalmology Cohort} For glaucoma detection, we used the Harvard-Fair\-VLMed dataset \cite{FairCLIP}, containing 10,000 fundus photographs with age, gender, and race as selected attributes. Given the exponential increase in intersectional subgroups when three attributes are combined, we adopted a binary age split (0-60 vs.\ 60+) and a binary race split (White vs.\ Non-White), yielding eight intersectional subgroups. The age threshold of 60 years is strongly justified in ophthalmology, as it marks an exponential increase in glaucoma prevalence from approximately 1\% to over 3\% \cite{Zhang2021glaucoma}. Race binarisation, while a simplification, was a pragmatic decision driven by the dataset's distribution to ensure all eight subgroups met minimum sample size requirements for robust analysis; further subdivision would have created low-count subgroups ($n < 50$) that compromise statistical validity \cite{Yang2024limits}. The dataset was split into training ($n = 5{,}968$; 60\%), validation ($n = 2{,}000$; 20\%), and held-out test ($n = 2{,}032$; 20\%) sets with stratified sampling.

\textbf{Data Quality and Pairing} All images were verified to have ground-truth diagnostic labels confirmed by histopathology (HAM10000, BCN20000) or clinical assessment (Harvard-FairVLMed). For VLM training, each image was paired with a text description: for skin lesions, structured sentences embedding the disease label (e.g., ``A dermoscopic image showing a benign melanocytic nevus''); for fundus images, clinical note summaries derived from the original reports. Test sets were strictly held out throughout model development and used exclusively for final performance evaluation. Demographic attributes were used only during training to compute the fairness regularisation term and were never provided as model inputs during inference, preserving patient privacy.

\subsection{Quantifying the Diagnostic Certainty Gap}\label{subsec2}

We posit that even when models achieve similar aggregate accuracy, they may exhibit profound differences in the confidence of their predictions, leaving marginalised subgroups in a zone of uncertainty where diagnoses become unstable and vulnerable to misclassification from minor data perturbations. To formalise and quantify this phenomenon, we defined a per-sample diagnostic certainty score and analysed its distribution across intersectional subgroups.

\textbf{Diagnostic Certainty Score Definition} For a vision-language model producing normalised image embeddings $\mathbf{z}^I$ and text embeddings $\mathbf{z}^T$, we defined the diagnostic certainty for sample $i$ as the softmax-calibrated probability assigned to the correct diagnostic class:
\begin{equation}
c_i = \frac{\exp\left(\langle \mathbf{z}_i^I, \mathbf{z}_{\text{correct}}^T \rangle / \tau\right)}{\sum_{k \in \mathcal{C}} \exp\left(\langle \mathbf{z}_i^I, \mathbf{z}_k^T \rangle / \tau\right)},
\end{equation}
where $\langle \cdot, \cdot \rangle$ denotes cosine similarity, $\mathcal{C}$ is the set of candidate diagnostic classes (e.g., malignant vs.\ benign for skin lesions; glaucoma vs.\ non-glaucoma for fundus images), and $\tau$ is the temperature parameter. This score ranges from 0 to 1, with values near 0.5 indicating maximal uncertainty at the decision boundary. Predictions with certainty scores clustered near the decision threshold are clinically problematic: they are susceptible to reversal under minor variations in imaging conditions, patient positioning, or acquisition device, eroding clinician trust and producing inconsistent diagnostic recommendations.

\textbf{Distributional Analysis Across Subgroups} To characterise certainty disparities, we computed the distribution of diagnostic certainty scores separately for each intersectional subgroup in both pretrained and fine-tuned models. We employed Kernel Density Estimation (KDE) with a Gaussian kernel and a bandwidth selected to generate smooth probability density estimates, enabling visualisation of the full distributional shape rather than summary statistics alone. We defined the zone of uncertainty as the interval $[0.40, 0.60]$ surrounding the decision threshold based on statistical analysis, and quantified the proportion of each subgroup's predictions that fall within this zone.

\textbf{Certainty Gap Metric} We formalised the intersectional certainty gap as the maximum difference in mean diagnostic certainty between any two subgroups:
\begin{equation}
\Delta_{\text{certainty}} = \max_{g, g' \in \mathcal{G}} \left| \mathbb{E}[c \mid g] - \mathbb{E}[c \mid g'] \right|,
\end{equation}
where $\mathcal{G}$ denotes the set of intersectional subgroups. This metric assesses whether certain patient populations systematically receive less definitive diagnostic outputs, regardless of whether final classification accuracy appears equitable.

\subsection{The Cross-Modal Alignment Consistency (CMAC-MMD) Framework}\label{subsec3}

%% Figure 2
\begin{figure}[ht]
\centering
\includegraphics[width=\textwidth]{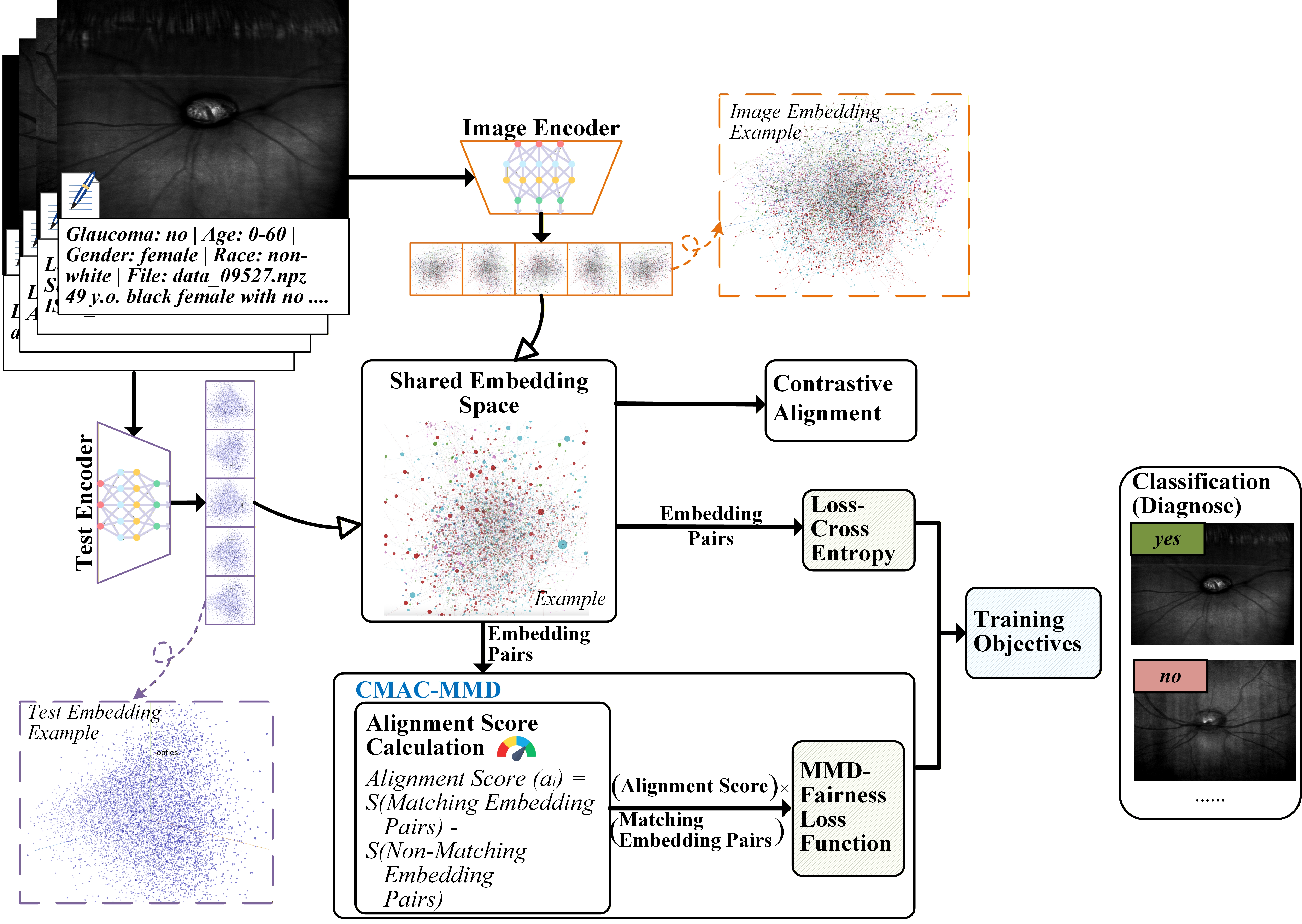}
\caption{\textbf{An overview of the proposed CMAC-MMD method with calculation of the Diagnostic Certainty Score and its role in fairness optimisation.} The schematic illustrates how the model quantifies diagnostic confidence for individual patient samples. The Image and Text Encoders map clinical inputs into a shared embedding space. The Alignment Score Calculation (ASC) derives a scalar value ($a_i$) representing the margin of safety. Specifically, the difference between the model's similarity to the correct diagnosis (Matching Embedding Pair) versus the most compelling incorrect alternative (Non-Matching Embedding Pair). A higher score indicates a decisive diagnosis, while a score near zero represents the grey zone of uncertainty. These certainty scores are the direct inputs for the CMAC-MMD Fairness Loss Function, which penalises the model only when the distribution of these certainty scores differs between demographic subgroups, ensuring that marginalised populations are not systematically subjected to lower diagnostic confidence.}\label{fig2}
\end{figure}

We developed CMAC-MMD, a training framework that directly regularises diagnostic certainty across intersectional patient subgroups (overview in Fig.~\ref{fig2}). Unlike conventional fairness interventions that operate on high-dimensional feature representations, CMAC-MMD targets the model's decision-level outputs, ensuring that diagnostic confidence is equally reliable regardless of patient demographics.

\textbf{Base Architecture} Our method builds upon the Contrastive Language-Image Pre-training (CLIP) framework \cite{CLIP}, which learns a shared embedding space for images and text through contrastive learning. Let the dataset be $\mathcal{D} = \{(\mathbf{I}_n, \mathbf{T}_n, y_n, \mathbf{a}_n)\}_{n=1}^N$, where each sample comprises an image $\mathbf{I}_n$, paired text description $\mathbf{T}_n$, disease label $y_n$, and demographic attributes $\mathbf{a}_n$. The image encoder $\phi_\theta(\cdot)$ and text encoder $\psi_\phi(\cdot)$ produce $\ell_2$-normalised embeddings $\mathbf{z}^I$ and $\mathbf{z}^T$, respectively. The standard training objective maximises cosine similarity between matched image-text pairs while minimising similarity for mismatched pairs.

\textbf{Alignment Score as Diagnostic Certainty} We define a scalar alignment score $a_i$ for each sample that quantifies the model's diagnostic decisiveness. This score measures the margin by which the model prefers the correct diagnosis over its most compelling alternative:
\begin{equation}
a_i = S_{ii} - \max_{j \neq i} S_{ij},
\end{equation}
where $S_{ij} = \langle \mathbf{z}_i^I, \mathbf{z}_j^T \rangle / \tau$ denotes the temperature-scaled cosine similarity. A positive score ($a_i > 0$) indicates confident separation between correct and incorrect diagnoses; scores near zero indicate borderline predictions where the model cannot decisively distinguish diagnostic alternatives. Clinically, this score directly reflects whether a patient's diagnosis falls within a reliable range or remains in an uncertain grey zone susceptible to misclassification.

%% Figure 3
\begin{figure}[ht]
\centering
\includegraphics[width=\textwidth]{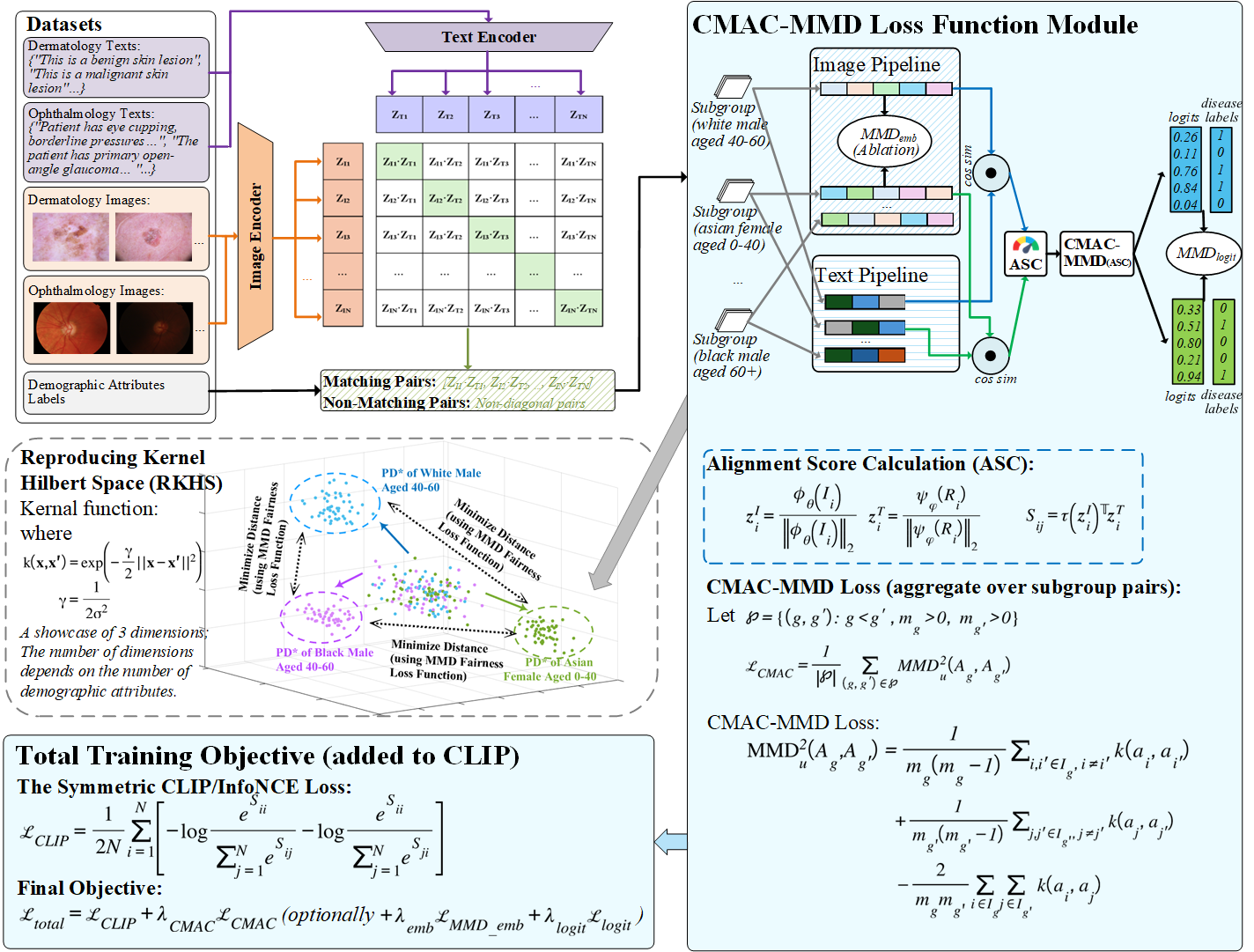}
\caption{\textbf{The detailed CMAC-MMD training framework.} The model takes image-text pairs and demographic attributes as input. Instead of regularising high-dimensional embeddings, CMAC-MMD first computes a per-sample scalar ASC that quantifies the model's diagnostic certainty. These scalar scores are grouped by subgroup, and the CMAC-MMD loss is calculated to result in 1-D distributions, enforcing that the distribution of model certainty is consistent across all intersectional patient groups. The diagram also illustrates the placement of the ablation MMD regularizers, applied directly to the image embeddings ($\text{MMD}_{\text{emb}}$) and final disease logits ($\text{MMD}_{\text{logit}}$) for comparative analysis. The Reproducing Kernel Hilbert Space (RKHS) visualisation and radial basis function (RBF) kernel \cite{ghojogh2021reproducing, thurnhofer2020radial} depict the conceptual goal: minimising the distance between the probability distributions ($\text{PD}^*$) of different subgroups to achieve fairness. The final training objective combines the standard symmetric Contrastive Language-Image Pre-training (CLIP)/Information Noise-Contrastive Estimation (InfoNCE) loss with the weighted $\mathcal{L}_{\text{CMAC}}$ penalty.}\label{fig3}
\end{figure}

\textbf{Distributional Fairness via Maximum Mean Discrepancy} Our core hypothesis is that for a model to be fair, the entire distribution of alignment scores should be consistent across all intersectional subgroups, not merely their mean values. Two subgroups may exhibit identical average diagnostic performance yet differ substantially in reliability: one receiving consistently confident predictions while another experiences a disproportionate share of uncertain, borderline diagnoses. We enforce distributional consistency using the Maximum Mean Discrepancy (MMD) \cite{gretton2012kernel}, a kernel-based statistical test that measures the distance between probability distributions. For each subgroup $g$ present in a mini-batch, we form the distribution of its alignment scores $A_g = \{a_i \mid \mathbf{a}_i \in g\}$. The CMAC-MMD loss is computed as:
\begin{equation}
\mathcal{L}_{\text{CMAC}} = \frac{1}{|\mathcal{P}|} \sum_{(g, g') \in \mathcal{P}} \text{MMD}^2(A_g, A_{g'}),
\end{equation}
where $\mathcal{P}$ denotes all pairwise combinations of subgroups in the batch, and MMD is computed using a radial basis function kernel. By minimising this loss, the model is explicitly trained to produce statistically indistinguishable certainty distributions across all demographic intersections.

\textbf{Total Training Objective} The final objective combines the standard contrastive loss with the fairness regularisation term:
\begin{equation}
\mathcal{L}_{\text{Total}} = \mathcal{L}_{\text{CLIP}} + \lambda_{\text{CMAC}} \mathcal{L}_{\text{CMAC}},
\end{equation}
where the hyperparameter $\lambda_{\text{CMAC}}$ controls the strength of fairness regularisation, enabling practitioners to calibrate the trade-off between overall diagnostic performance and intersectional equity based on clinical priorities (Fig.~\ref{fig3}). A critical design feature of CMAC-MMD is that, during training, demographic attributes are used exclusively to compute the fairness loss. During clinical inference, the model requires only the medical image and does not access patient demographic information, preserving privacy and enabling deployment in settings where such data may be unavailable or protected by regulation. Complete gradient derivations for the CMAC-MMD loss function, including theoretical fairness guarantees, are provided (Supplementary Appendix D).

\subsection{Experimental Design and Baseline Comparisons}\label{subsec4}

We designed a comprehensive experimental framework comparing our method against established fairness interventions spanning multiple algorithmic paradigms. We evaluated intersectional fairness across a diverse suite of VLMs to ensure generalisability of our findings. The primary experiments used CLIP with a Vision Transformer (ViT-B/16) backbone \cite{CLIP}, which serves as the foundational architecture for contrastive vision-language learning. We additionally assessed biomedical-adapted variants, including BioMedCLIP \cite{BioMedCLIP}, PMC-CLIP \cite{PMC-CLIP}, PubMedCLIP \cite{PubMedCLIP}, and MedCLIP \cite{MedCLIP}, which were pre-trained on medical image-text corpora. For the ophthalmology experiments comparing against FairCLIP \cite{FairCLIP}, we used identical architectural configurations to ensure fair comparison.

Before evaluating fairness interventions, we conducted a preliminary experiment to empirically characterise the relationship between standard fine-tuning and intersectional bias. Each VLM architecture was fine-tuned on the dermatology dataset using standard ERM without fairness constraints. We compared classification performance and fairness metrics between pretrained and fine-tuned states across all intersectional subgroups. This experiment tests the hypothesis that domain adaptation, while improving overall diagnostic accuracy, systematically exacerbates performance disparities for marginalised subgroups, thereby establishing the clinical need for fairness-aware training methods such as CMAC-MMD.

We compared CMAC-MMD against seven established fairness interventions representing three distinct algorithmic categories:

\begin{itemize}
    \item \textit{Standard training (ERM):} Empirical Risk Minimisation (ERM) without fairness constraints, serving as the reference baseline.
    \item \textit{Data-level pre-processing:} Resampling \cite{Resam}, which balances training data by oversampling minority subgroups; and Reweighting \cite{Rewei}, which adjusts sample importance based on subgroup membership.
    \item \textit{Algorithmic in-processing:} Group Distributionally Robust Optimisation (GroupDRO) \cite{DRO}, which optimises worst-group performance; Mean Accuracy, which explicitly balances per-group accuracy during training; Domain-Adversarial Neural Networks (DANN) \cite{DANN} and Conditional DANN (CDANN) \cite{CDANN}, which learn subgroup-invariant representations through adversarial training.
    \item \textit{VLM-specific fairness:} FairCLIP \cite{FairCLIP}, a recent method explicitly designed for vision-language models that applies fairness constraints to individual demographic attributes.
\end{itemize}

This selection encompasses the current state-of-the-art across data-centric, representation-learning, and VLM-specific fairness paradigms, enabling a comprehensive assessment of CMAC-MMD's relative effectiveness. Detailed algorithmic adaptations for VLM architectures are provided (Supplementary Appendix C).

All models were fine-tuned for 20 epochs (dermatology) and 50 epochs (ophthalmology) using the Adam with weight decay (AdamW) optimiser with learning rate $1 \times 10^{-5}$ and weight decay $5 \times 10^{-5}$ (complete hyperparameter configuration provided in Supplementary Table E9). For CMAC-MMD, the fairness regularisation strength was set to $\lambda_{\text{CMAC}} = 0.5$ based on validation set performance. Batch sizes were configured to ensure adequate representation of intersectional subgroups within each mini-batch, a requirement for computing the MMD-based fairness loss. All experiments were repeated three times with different random seeds to assess variability; we report mean performance with 95\% confidence intervals. Models were trained on two NVIDIA RTX 4090 GPUs; complete implementation details and code availability will be provided in GitHub repository upon acceptance. For dermatology, models were trained on HAM10000 and evaluated on both the held-out HAM10000 test set (internal validation) and the independent BCN20000 dataset (external validation) to assess generalisation under distribution shift. For ophthalmology, models were trained and evaluated on Harvard-FairVLMed with stratified held-out test sets. All reported metrics are computed on strictly held-out test data that were not used during model development or hyperparameter selection.

\subsection{Statistical Analysis and Performance Metrics}\label{subsec5}

We pre-specified primary and secondary outcome measures prior to conducting experiments, following reporting standards for clinical AI evaluation studies \cite{Liu2025fairness}. All statistical analyses were performed using Python 3.12.3 with SciPy 1.11 and NumPy 1.24.

We designated two co-primary endpoints to jointly assess the study aims:

\begin{itemize}
    \item \textit{Diagnostic Performance:} Area Under the Receiver Operating Characteristic Curve (AUC), measuring the model's overall discriminative ability to distinguish diseased from non-diseased cases across all operating thresholds.
    \item \textit{Fairness Performance:} Difference in True Positive Rate ($\Delta$TPR), defined as the maximum disparity in sensitivity between any two intersectional subgroups:
    \begin{equation}
    \Delta\text{TPR} = \max_{g, g' \in \mathcal{G}} \left| \text{TPR}(g) - \text{TPR}(g') \right|,
    \end{equation}
    where $\text{TPR}(g) = \text{TP}_g / (\text{TP}_g + \text{FN}_g)$ for subgroup $g$. We selected $\Delta$TPR as the primary fairness metric because in clinical screening, a disparity in true positive rate directly represents a disparity in missed diagnoses, which carries the most consequential impact on patient safety---a missed melanoma or undetected glaucoma translates to delayed treatment and poorer prognosis.
\end{itemize}

To provide a comprehensive fairness assessment, we evaluated additional metrics with direct clinical interpretations:

\begin{itemize}
    \item \textit{Demographic Parity Difference (DPD):} The maximum disparity in positive prediction rates across subgroups, defined as $\text{DPD} = \max_{g, g'} | P(\hat{Y}=1 \mid g) - P(\hat{Y}=1 \mid g') |$. This metric assesses whether the model recommends further clinical action (e.g., biopsy, specialist referral) at equitable rates across patient populations.
    \item \textit{Difference in False Positive Rate ($\Delta$FPR):} The maximum disparity in false alarm rates, representing inequitable exposure to unnecessary procedures or patient anxiety.
    \item \textit{Difference in Equalised Odds (DEOdds):} A composite metric capturing disparities in both sensitivity and specificity, computed per subgroup as $\text{DEOdds}(g) = |\text{TPR}(g) - \overline{\text{TPR}}| + |\text{FPR}(g) - \overline{\text{FPR}}|$, where $\overline{\text{TPR}}$ and $\overline{\text{FPR}}$ denote population-level rates.
\end{itemize}

Beyond continuous metrics, we evaluated binary fairness criteria specifically designed for intersectional analysis \cite{Foulds2020, Gaurav2023}:

\begin{itemize}
    \item \textit{Differential Fairness (DF):} A method satisfies DF at level $\varepsilon$ if the ratio of true positive rates between any two subgroups is bounded: $e^{-\varepsilon} \leq \text{TPR}(g) / \text{TPR}(g') \leq e^{\varepsilon}$ for all $g, g' \in \mathcal{G}$. We adopted $\varepsilon = 0.5$ as the threshold, corresponding to a maximum 1.65-fold ratio in detection rates.
    \item \textit{Intersectional Fairness-$\alpha$ (IF-$\alpha$):} A criterion that guards against ``levelling down'' by jointly penalising absolute and relative performance disparities: $L_\alpha(g, g') = \alpha \cdot \Delta_{\text{abs}} + (1-\alpha) \cdot \Delta_{\text{rel}}$. We used $\alpha = 0.5$ and threshold $\gamma = 0.4$, ensuring fairness is achieved by improving outcomes for disadvantaged groups rather than degrading performance universally.
\end{itemize}

We employed the following pre-specified statistical tests to determine whether observed differences were statistically significant:

\begin{itemize}
    \item \textit{DeLong Test} \cite{DeLong1988} for comparing paired AUC values between CMAC-MMD and each baseline method. This test accounts for the correlation between AUCs computed on the same test set and provides asymptotically valid confidence intervals. We report two-sided $p$-values and 95\% confidence intervals for AUC differences.
    \item \textit{Wilcoxon Signed-Rank Test} \cite{Wilcoxon1945} for comparing paired distributions of subgroup-level fairness metrics (DEOdds) between methods. This non-parametric test is appropriate for comparing matched observations across intersectional subgroups without distributional assumptions.
    \item \textit{Two-proportion Z-test} for comparing aggregate fairness metrics (DPD, $\Delta$TPR) between methods, using normal approximation for large samples.
    \item \textit{Bootstrap Confidence Intervals:} We employed stratified bootstrapping with 10,000 resamples to generate 95\% percentile confidence intervals for all reported metrics, ensuring robust uncertainty quantification.
\end{itemize}

Statistical significance was defined as $p < 0.05$ (two-sided). We applied no correction for multiple comparisons across baseline methods, as each comparison addresses a distinct scientific question regarding CMAC-MMD's relative performance; however, we report exact $p$-values to enable reader interpretation.

To translate statistical improvements into clinically meaningful terms, we quantified the potential reduction in missed diagnoses attributable to CMAC-MMD. For each intersectional subgroup $g$, we calculated:
\begin{equation}
\text{FN}_{\text{prevented}}(g) = N_g \times \pi_g \times \left[ (1 - \text{TPR}_{\text{baseline}}(g)) - (1 - \text{TPR}_{\text{CMAC}}(g)) \right],
\end{equation}
where $N_g$ is the subgroup sample size, $\pi_g$ is the disease prevalence within that subgroup, and the bracketed term represents the reduction in false negative rate. This calculation projects the number of patients within each demographic intersection who would receive a correct positive diagnosis under CMAC-MMD but would be missed under baseline approaches. We report both absolute counts and relative reductions to contextualise the clinical significance of observed improvements. All experiments were conducted three times with different random initialisations; we report mean values with 95\% confidence intervals derived from bootstrap resampling. Consistent with the study aims, we declare CMAC-MMD successful if it demonstrates: (1) non-inferior or superior AUC compared to baseline ($\Delta$AUC $\geq -0.02$), AND (2) statistically significant reduction in $\Delta$TPR ($p < 0.05$). Results are reported separately for each clinical domain (dermatology, ophthalmology) and validation setting (internal, external) to assess consistency across datasets and scenarios. A summary of the evaluation protocol is also provided (Supplementary Appendix E.2).

\section{Results}\label{sec3}

\subsection{Cohort Characteristics and the Source of Diagnostic Disparities}\label{subsec6}
\begin{table}[h]
\caption{\textbf{Dataset selection for intersectional fairness analysis.} Datasets were evaluated based on sample size, availability of multiple demographic attributes for intersectional analysis, and label verification method.}\label{table1}%
\begin{tabular*}{\textwidth}{@{\extracolsep\fill}lccccc@{}}
\toprule
Dataset & \#Images & \#Patients & Verification & Attributes & Selected \\
\midrule
\multicolumn{6}{@{}c@{}}{\textit{Dermatology (Skin Lesion)}} \\
\midrule
HAM10000 \cite{DVN/DBW86T_2018} & 10,015 & NR & HP/CM/CF/EC & Age, Gender & $\checkmark$ \\
BCN20000 \cite{BCN} & 18,946 & 5,583 & HP & Age, Gender & $\checkmark$ \\
Fitzpatrick17k \cite{groh2021evaluating} & 16,577 & NR & Unverified\footnotemark[1] & Skin Type & $\times$ \\
PAD-UFES-20 \cite{PACHECO2020106221} & 2,298 & 1,373 & HP (58\%) & Age, Gender & $\times$ \\
DDI \cite{DDI} & 656 & 570 & HP & Skin Type & $\times$ \\
\midrule
\multicolumn{6}{@{}c@{}}{\textit{Ophthalmology (Glaucoma)}} \\
\midrule
Harvard-FairVLMed \cite{FairCLIP} & 10,000 & 10,000 & VF/Clinical & Age, Gender, Race & $\checkmark$ \\
LAG \cite{Li_2019_CVPR} & 5,824 & NR & Clinical & Limited & $\times$ \\
PAPILA \cite{Kovalyk2022} & 488 & NR & Clinical & Age, Gender & $\times$ \\
ACRIMA \cite{ovreiu2021deep} & 705 & NR & Clinical & Limited & $\times$ \\
ORIGA \cite{5626137} & $\sim$650 & NR & Clinical & Unknown & $\times$ \\
\botrule
\end{tabular*}
\footnotetext{Note: NR, not reported. HP, histopathology; CM, confocal microscopy; CF, clinical follow-up; EC, expert consensus; VF, visual field test.}
\footnotetext[1]{Expert review of 3\% sample found only 69\% clearly diagnostic of labeled condition \cite{abhishek2025investigating}.}
\end{table}

We included HAM10000 \cite{DVN/DBW86T_2018}, BCN20000 \cite{BCN}, and Harvard-FairVLMed \cite{FairCLIP} as datasets for this study, after evaluating a broad set of benchmark datasets in dermatology and ophthalmology (Table~\ref{table1}). Other commonly used datasets were determined to be unsuitable for rigorous intersectional fairness analysis because they either lacked sufficient demographic attributes to construct intersectional subgroups or contained insufficient sample sizes within resulting subgroups to support statistically valid metric estimation. Specifically, Fitzpatrick17k \cite{groh2021evaluating} provides only Fitzpatrick skin type without age or gender; PAD-UFES-20 \cite{PACHECO2020106221} and DDI \cite{DDI} contain fewer than 2,300 and 700 images respectively, yielding subgroup counts below the minimum threshold of $n = 100$ required for reliable fairness evaluation \cite{RicciLara2022}. Similarly, ophthalmology benchmarks including LAG, PAPILA, ACRIMA, and ORIGA lacked the demographic metadata necessary for intersectional analysis.

%% Figure 4: Certainty gap illustration
\begin{figure}[ht]
\centering
\includegraphics[width=\textwidth]{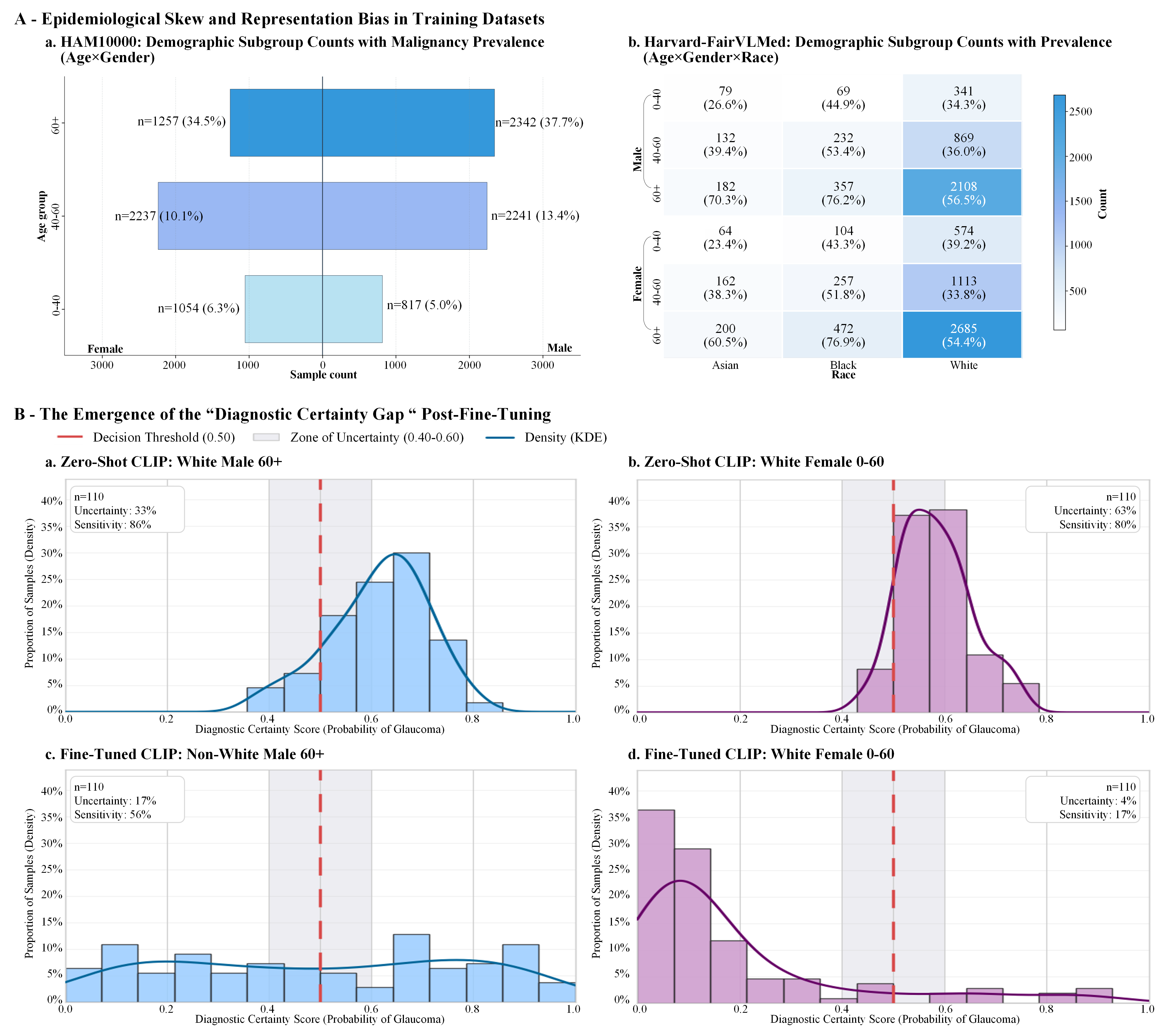}
\caption{\textbf{Demographic data imbalance drives systematic disparities in diagnostic certainty.} \textbf{A} Distribution of intersectional subgroups in the dermatology and ophthalmology datasets. Bar charts and heatmaps reveal significant population imbalances, with specific intersections overrepresented relative to underrepresented subgroups. \textbf{B} KDE plots visualise the distribution of the model’s predicted probabilities (diagnostic confidence) for intersectional subgroups. The ``Grey Zone'' (shaded) represents predictions near the decision threshold (0.5) that are uncertain.
}\label{fig4}
\end{figure}

\textbf{Demographic Imbalance in Training Data} The included datasets exhibit substantial representation imbalances that provide context for the observed diagnostic disparities (Fig.~\ref{fig4}A). In the HAM10000 dermatology cohort, males aged 60+ constituted the largest subgroup ($n = 2{,}342$; 23.4\% of the dataset) with the highest malignancy prevalence (37.7\%), while females aged 0-40 represented a smaller proportion ($n = 1{,}054$; 10.5\%) with substantially lower disease prevalence (6.3\%). This 6-fold difference in malignancy prevalence across age-gender subgroups creates a learning environment where models are exposed to vastly different numbers of positive cases per subgroup. The Harvard-FairVLMed ophthalmology cohort demonstrated even more pronounced imbalances across the three-attribute intersection of age, gender, and race. White patients aged 60+ dominated the dataset ($n = 2{,}685$ for females and $n = 2{,}108$ for males), while non-white patients aged 0-60 were substantially underrepresented ($n = 257$ for females and $n = 162$ for males). Disease prevalence also varied across subgroups, ranging from 26.6\% in young Asian males to 76.9\% in older Black males.

\textbf{Standard Fine-Tuning Creates a Diagnostic Certainty Gap} Beyond representation imbalance, we identified a systematic disparity in diagnostic certainty that emerges during standard model fairness-unaware fine-tuning, a phenomenon we term the ``diagnostic certainty gap'' (Fig.~\ref{fig4}B). To characterise this gap, we analysed the distribution of model confidence scores across intersectional subgroups before and after fine-tuning using Kernel Density Estimation (KDE). We defined the zone of uncertainty as the interval [0.40, 0.60] surrounding the decision threshold, where predictions are clinically unreliable and susceptible to reversal under minor data perturbations. In the zero-shot (pretrained) CLIP model evaluated on the ophthalmology dataset, both the majority subgroup (White Male 60+; $n = 110$) and an underrepresented subgroup (White Female 0-60; $n = 110$) exhibited predictions within the zone of uncertainty at rates of 33\% and 63\%, respectively, with corresponding sensitivities of 86\% and 80\%. After standard fine-tuning, the model's behaviour diverged dramatically between subgroups. For the Non-White Male 60+ subgroup, fine-tuning reduced uncertain predictions to 17\% while maintaining clinically acceptable sensitivity (56\%). In stark contrast, for the White Female 0-60 subgroup, although the proportion of uncertain predictions decreased to just 4\%, this apparent improvement masked a catastrophic collapse in diagnostic performance: sensitivity plummeted from 80\% to 17\%. The model had learned to predict this subgroup as predominantly negative with high confidence, a pattern that would result in the systematic underdiagnosis of glaucoma in young white female patients. Specifically, 83\% of glaucoma cases in this subgroup would be missed, compared to 44\% in the better-represented male subgroup.

%% Figure: Fine-tuning degradation across architectures
\begin{figure}[!ht]
\centering
\includegraphics[width=\textwidth]{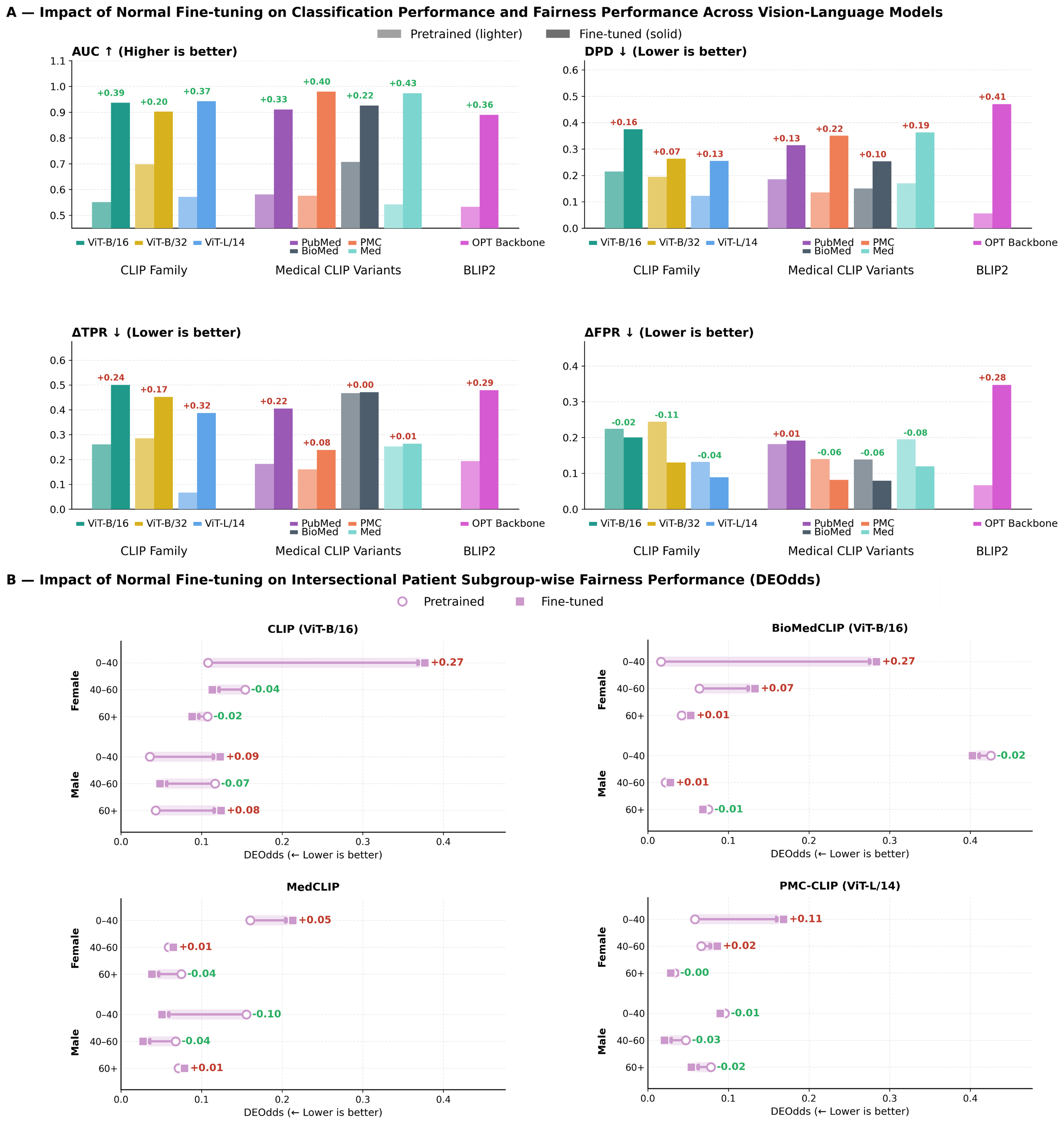}
\caption{\textbf{Standard fine-tuning improves overall accuracy but systematically degrades intersectional fairness across vision-language model architectures.} \textbf{A} Comparison of pretrained (light bars) versus fine-tuned (solid bars) performance across eight VLM architectures. Numbers above bars indicate the change from pretrained to fine-tuned state. \textbf{B} Subgroup-level DEOdds for four representative models. Dumbbell plots show the shift from pretrained (hollow circles) to fine-tuned (solid squares) states for each intersectional subgroup. Rightward movement indicates worsening fairness.}
\label{fig5}
\end{figure}

\textbf{Fine-Tuning Degradation of Fairness is Systematic Across Vision-Language Architectures} To assess whether this fairness-accuracy trade-off generalises beyond a single model, we evaluated eight VLM architectures spanning three model families: the CLIP family (ViT-B/16, ViT-B/32, ViT-L/14), medical domain-adapted variants (PubMedCLIP, BioMedCLIP, PMC-CLIP, MedCLIP), and BLIP2 (Fig.~\ref{fig5}). Standard fine-tuning improved overall AUC across all architectures, with gains ranging from +0.20 (CLIP ViT-B/32) to +0.43 (MedCLIP). However, this improvement was consistently accompanied by degradation in fairness metrics. The disparity in true positive rates ($\Delta$TPR) increased in seven of eight models, with the largest deterioration observed in CLIP ViT-B/16 (+0.24) and BLIP2 (+0.29). Subgroup-level analysis confirmed that fairness degradation disproportionately affected specific intersectional groups: across all four models evaluated at the subgroup level, Female 0-40 patients exhibited the largest increase in DEOdds after fine-tuning, with deterioration ranging from +0.05 (MedCLIP) to +0.27 (CLIP ViT-B/16 and BioMedCLIP). These findings demonstrate that the fairness-accuracy trade-off is intrinsic to standard fine-tuning rather than an artefact of a particular architecture, underscoring the need for fairness-aware training methods such as CMAC-MMD.

\subsection{CMAC-MMD Improves Diagnostic Performance While Reducing Missed Diagnosis Disparities in Dermatology}\label{subsec7}

\begin{table}[ht]
\caption{\textbf{Comparison of fairness interventions for skin lesion classification (HAM10000).} CMAC-MMD achieves the highest AUC while simultaneously producing the lowest disparity metrics. $\Delta$TPR represents the maximum difference in true positive rate (sensitivity) between any two intersectional subgroups. Clinically, this quantifies the gap in missed diagnosis rates.}
\label{table2}
\small
\begin{tabular*}{\textwidth}{@{\extracolsep\fill}lcccccccc@{}}
\toprule
& \multicolumn{3}{@{}c@{}}{Diagnostic Performance} & \multicolumn{3}{@{}c@{}}{Fairness Metrics} & \multicolumn{2}{@{}c@{}}{Criteria} \\
\cmidrule{2-4} \cmidrule{5-7} \cmidrule{8-9}%
Method & AUC$\uparrow$ & $p$\footnotemark[1] & DPD$\downarrow$ & $\Delta$TPR$\downarrow$ & DEOdds$\downarrow$ & DF & IF-$\alpha$ \\
\midrule
ERM (Baseline) & 0.94 & $<$0.001 & 0.38 & 0.50 & 0.146 & $\times$ & $\times$ \\
Resampling \cite{Resam} & 0.96 & $<$0.05 & 0.44 & 0.31 & 0.106 & $\checkmark$ & $\checkmark$ \\
Reweighting \cite{Rewei} & 0.97 & 0.56 & 0.36 & 0.28 & 0.081 & $\times$ & $\times$ \\
Mean Accuracy & 0.92 & $<$0.001 & 0.43 & 0.31 & 0.116 & $\times$ & $\times$ \\
GroupDRO \cite{DRO} & 0.92 & $<$0.001 & 0.41 & 0.46 & 0.159 & $\times$ & $\times$ \\
DANN \cite{DANN} & 0.96 & $<$0.05 & 0.31 & 0.42 & 0.149 & $\times$ & $\times$ \\
CDANN \cite{CDANN} & 0.97 & $<$0.001 & 0.37 & 0.27 & 0.115 & $\checkmark$ & $\checkmark$ \\
\textbf{CMAC-MMD} & \textbf{0.97} & ref. & \textbf{0.30} & \textbf{0.26} & \textbf{0.058} & $\checkmark$ & $\checkmark$ \\
\botrule
\end{tabular*}
\footnotetext[1]{Two-sided DeLong test $p$-value for AUC comparison versus CMAC-MMD (reference).}
\end{table}

We evaluated CMAC-MMD against a standard ERM baseline and seven established fairness interventions on the HAM10000 dermatology cohort (Table~\ref{table2}). CMAC-MMD achieved the highest overall diagnostic performance (AUC = 0.97; 95\% confidence interval (CI): 0.96-0.98), significantly outperforming the ERM baseline (AUC = 0.94; $\Delta$AUC = +0.03; 95\% CI: 0.030-0.063; two-sided DeLong $p < 0.0001$). This improvement was comparable to Reweighting and CDANN (both AUC = 0.97), while substantially exceeding GroupDRO (AUC = 0.92; $\Delta$AUC = +0.05; $p < 0.0001$) and Mean Accuracy (AUC = 0.92; $\Delta$AUC = +0.06; $p < 0.0001$). Concurrently, CMAC-MMD achieved the greatest reduction in diagnostic disparities across intersectional subgroups. The maximum gap in true positive rate ($\Delta$TPR), which directly quantifies the disparity in missed diagnoses between the best and worst-performing subgroups, decreased from 0.50 under ERM to 0.26 under CMAC-MMD, a 48\% relative reduction ($z = 16.10$, two-sided $p < 0.0001$). Similarly, DPD decreased from 0.38 to 0.30 ($z = 5.35$, $p < 0.0001$). The mean Difference in DEOdds across all six intersectional subgroups was reduced from 0.146 (ERM) to 0.058 (CMAC-MMD), representing a 60\% improvement (Wilcoxon signed-rank $W = 1.0$, $p = 0.0625$). CMAC-MMD was one of only three methods to satisfy both pre-specified binary fairness criteria: Differential Fairness (DF at $\varepsilon = 0.5$) and Intersectional Fairness-$\alpha$ (IF-$\alpha$ at $\alpha = 0.5$, $\gamma = 0.4$). Complete subgroup-level AUC comparisons against all baselines are provided (Supplementary Table B1), with corresponding fairness metrics presented (Supplementary Table B2).

\begin{figure}[!ht]
\centering
\includegraphics[width=\textwidth]{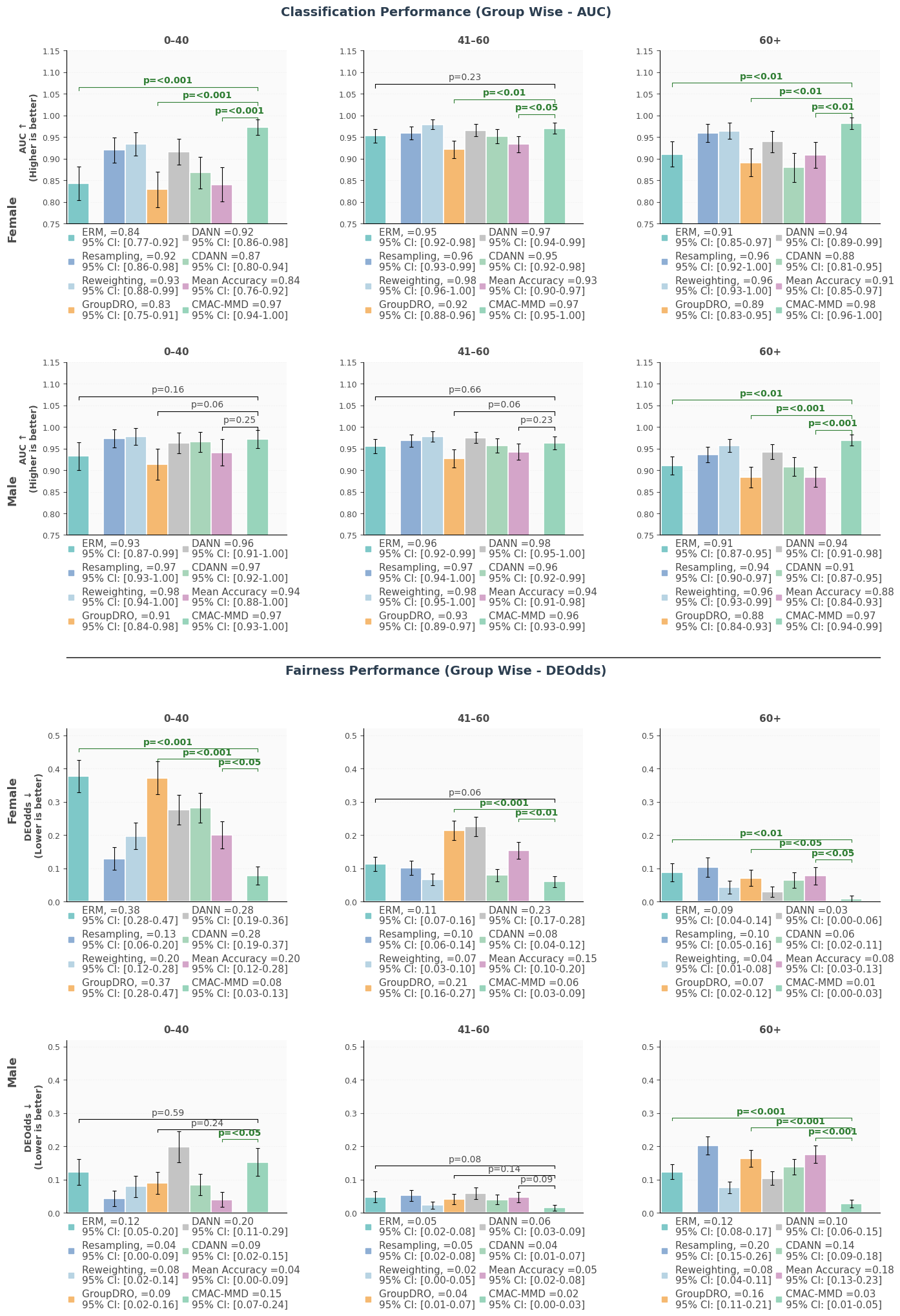}
\caption{\textbf{CMAC-MMD improves diagnostic performance and reduces disparities across all intersectional subgroups in dermatology.} Top two panels: AUC for female and male subgroups stratified by age. Bottom panels: DEOdds, where lower values indicate reduced disparity from population-level performance. Methods are grouped by intervention type: ERM baseline with no fairness constraints (teal), existing methods, and CMAC-MMD (hatched purple). Error bars indicate 95\% confidence intervals from bootstrap resampling ($n = 10{,}000$ iterations). Significance $p$ value labels reflect DeLong test comparisons against CMAC-MMD.}
\label{fig6}
\end{figure}

\textbf{Subgroup-Level Analysis Reveals Targeted Improvements for Vulnerable Populations} Granular analysis across the six intersectional subgroups revealed that CMAC-MMD produced consistent benefits, with the largest improvements observed in the subgroups most disadvantaged under baseline training (Fig.~\ref{fig6}). The Female 0--40 subgroup, which exhibited the worst baseline performance (ERM AUC = 0.84; 95\% CI: 0.77-0.92), achieved the greatest improvement under CMAC-MMD (AUC = 0.97; 95\% CI: 0.94-1.00; $\Delta$AUC = +0.13; $z = 3.81$, $p < 0.001$). Similar statistically significant gains were observed for Female 60+ ($\Delta$AUC = +0.07; $z = 2.81$, $p < 0.01$) and Male 60+ ($\Delta$AUC = +0.06; $z = 3.21$, $p < 0.01$). Several baseline fairness interventions failed to address the most vulnerable subgroups or produced paradoxical harms. GroupDRO, designed to optimise worst-group performance, paradoxically degraded AUC for Female 0--40 to 0.83 (95\% CI: 0.75--0.91) while offering no meaningful fairness improvement (DEOdds = 0.37 vs. 0.38 for ERM). Embedding space analysis reveals that CMAC-MMD achieves superior disease separation while maintaining embedding uniformity across demographics (Supplementary Fig. A2--A5). DANN reduced overall disparity but at the cost of maintained underperformance for specific subgroups (Female 0-40 AUC = 0.92).

%% Figure reference for subgroup analysis

\begin{table}[ht]
\caption{\textbf{Clinical impact assessment: reduction in missed skin cancer diagnoses by intersectional subgroup.} False negatives (FN) represent malignant lesions incorrectly classified as benign}
\label{table3}
\begin{tabular*}{\textwidth}{@{\extracolsep\fill}lccccc@{}}
\toprule
Subgroup & $n$ (Test) & True Positives & FN (ERM) & FN (CMAC) & FN Prevented \\
\midrule
Female 0--40 & 222 & 13 & 5 & 2 & 3 (60.0\%) \\
Female 41--60 & 459 & 55 & 6 & 3 & 3 (50.0\%) \\
Female 60+ & 248 & 67 & 12 & 9 & 3 (25.0\%) \\
Male 0--40 & 166 & 8 & 3 & 2 & 1 (33.3\%) \\
Male 41--60 & 415 & 66 & 10 & 9 & 1 (10.0\%) \\
Male 60+ & 480 & 170 & 28 & 21 & 7 (25.0\%) \\
\midrule
\textbf{Total} & \textbf{1,990} & \textbf{379} & \textbf{64} & \textbf{46} & \textbf{18 (28.1\%)} \\
\botrule
\end{tabular*}
\footnotetext{Note: FN, false negatives; ERM, empirical risk minimisation; CMAC, Cross-Modal Alignment Consistency. Percentages in the final column indicate relative reduction in missed diagnoses. CMAC refers to CMAC-MMD}
\end{table}

\textbf{Quantification of Prevented Missed Diagnoses} To translate these statistical improvements into clinically interpretable terms, we quantified the reduction in false negative diagnoses across intersectional subgroups (Table~\ref{table3}). In the held-out test set ($n = 1{,}990$), the ERM baseline produced 64 false negative diagnoses (missed malignancies) across all subgroups. CMAC-MMD reduced this to 46 false negatives, preventing 18 missed diagnoses, a 28.1\% overall reduction in diagnostic failures. The impact was most pronounced in historically underserved subgroups: among Female 0-40 patients ($n = 222$; 13 true malignancies), CMAC-MMD correctly identified 3 cases that would have been missed by the baseline, representing a 60\% relative reduction in false negatives for this subgroup. Among Male 60+ patients ($n = 480$; 170 true malignancies), 7 additional cases were correctly identified, preventing 25\% of baseline false negatives.

\subsection{CMAC-MMD Demonstrates Cross-Domain Generalisability in Glaucoma Detection}\label{subsec8}

\begin{table}[ht]
\caption{\textbf{Comparison of fairness methods for glaucoma detection (Harvard-FairVLMed).} CMAC-MMD maintains diagnostic performance while achieving superior fairness compared to baseline and FairCLIP variants. FairCLIP-Race optimises fairness for race only; FairCLIP-All optimises for all attributes sequentially, considering age, gender, and race.}\label{table4}%
\small
\begin{tabular*}{\textwidth}{@{\extracolsep\fill}lccccccc@{}}
\toprule
& \multicolumn{3}{@{}c@{}}{Diagnostic Performance} & \multicolumn{4}{@{}c@{}}{Fairness Metrics} \\
\cmidrule{2-4} \cmidrule{5-8}%
Method & AUC & $z$\footnotemark[1] & $p$ & DPD & $\Delta$TPR & $\Delta$FPR & DEOdds \\
\midrule
ERM (Baseline) & 0.71 & 1.69 & 0.091 & 0.41 & 0.41 & 0.22 & 0.152 \\
\midrule
FairCLIP-Race \cite{FairCLIP} & 0.67 & 4.72 & $<$0.001 & 0.39 & 0.43 & 0.24 & 0.114 \\
FairCLIP-All \cite{FairCLIP} & 0.67 & 5.17 & $<$0.001 & 0.61 & 0.66 & 0.31 & 0.167 \\
\midrule
\textbf{CMAC-MMD} & \textbf{0.72} & ref. & ref & \textbf{0.28} & \textbf{0.31} & \textbf{0.19} & \textbf{0.096} \\
\botrule
\end{tabular*}
\footnotetext[1]{DeLong test $z$-statistic for AUC comparison versus CMAC-MMD (reference).}
\end{table}

%% Figure reference for ophthalmology subgroup analysis
\begin{figure}[!ht]
\centering
\includegraphics[width=\textwidth]{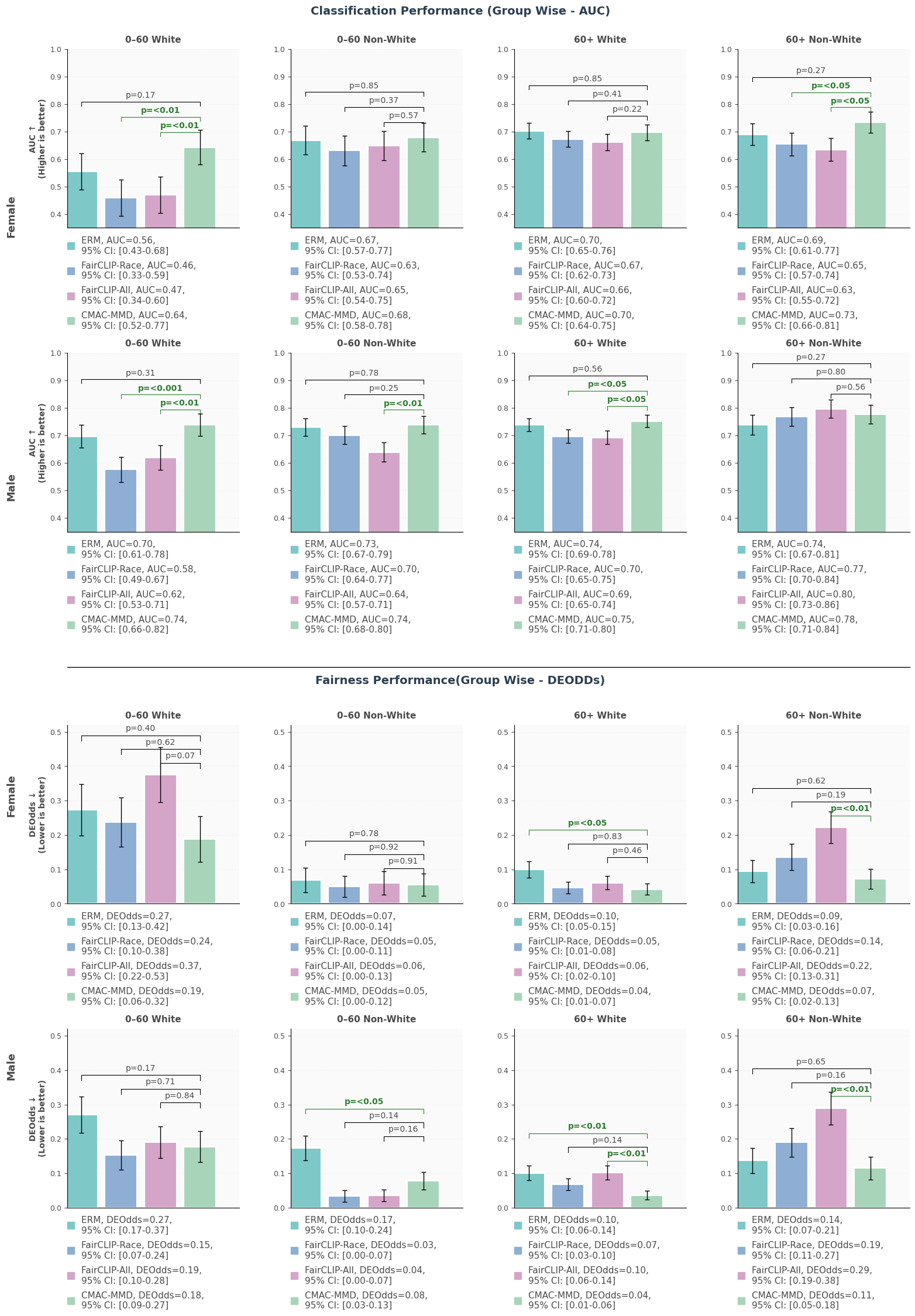}
\caption{\textbf{CMAC-MMD enhances performance and fairness in glaucoma detection across intersectional subgroups.} Top panels: AUC for female (left) and male (right) subgroups stratified by age and race (W, White; N-W, Non-White). Bottom panels: DEOdds metric. Error bars indicate 95\% confidence intervals. Significance labels reflect DeLong test comparisons ($p<0.05$).}
\label{fig7}
\end{figure}

To assess whether CMAC-MMD generalises beyond dermatology, we evaluated its performance on the Harvard-FairVLMed ophthalmology cohort for glaucoma detection, comparing against ERM and FairCLIP \cite{FairCLIP}, a fairness method specifically designed for vision-language models (Table~\ref{table4}). CMAC-MMD was the only method to simultaneously improve both diagnostic performance and fairness. It achieved an AUC of 0.72 (95\% CI: 0.70-0.74), representing a non-inferior improvement over the ERM baseline (AUC = 0.71; $\Delta$AUC = +0.01; 95\% CI: --0.003 to +0.045; two-sided DeLong $p = 0.09$). In contrast, both FairCLIP variants degraded overall diagnostic performance: FairCLIP-Race reduced AUC to 0.67 ($\Delta$AUC = -0.04 vs. ERM; $p < 0.001$) and FairCLIP-All similarly yielded AUC = 0.67 ($\Delta$AUC = -0.04; $p < 0.001$). CMAC-MMD significantly outperformed both FairCLIP variants (vs. FairCLIP-Race: $\Delta$AUC = +0.05, $p < 0.0001$; vs. FairCLIP-All: $\Delta$AUC = +0.05, $p < 0.0001$). Detailed subgroup-level comparisons are provided (Supplementary Table B3--B4).

CMAC-MMD achieved the most significant reduction in fairness disparities across all metrics. The maximum gap in true positive rate ($\Delta$TPR) decreased from 0.41 (ERM) to 0.31 (CMAC-MMD), a 24\% relative reduction ($z = 6.29$, $p < 0.0001$). DPD decreased from 0.41 to 0.28, representing a 32\% improvement ($z = 8.29$, $p < 0.0001$). Notably, FairCLIP-All, which attempts to optimise fairness across all demographic attributes simultaneously, paradoxically worsened both DPD (0.61) and $\Delta$TPR (0.66) compared to the baseline, illustrating the failure of single-attribute fairness optimisation when applied to intersectional subgroups. The mean DEOdds across all eight intersectional subgroups decreased from 0.152 (ERM) to 0.096 (CMAC-MMD), a 37\% improvement (Wilcoxon signed-rank $W = 0.0$, $p < 0.01$).

\textbf{Subgroup-Level Analysis Confirms Consistent Benefits Across Demographics} Analysis across the eight intersectional subgroups (age $\times$ gender $\times$ race) revealed that CMAC-MMD produced consistent improvements without the paradoxical harms observed with FairCLIP (Fig.~\ref{fig7}). CMAC-MMD significantly outperformed FairCLIP-Race in four subgroups: Female 0-60 White ($\Delta$AUC = +0.18; $z = 2.85$, $p < 0.01$), Male 0-60 White ($\Delta$AUC = +0.16; $z = 3.79$, $p < 0.001$), Female 60+ Non-White ($\Delta$AUC = +0.08; $z = 1.99$, $p < 0.05$), and Male 60+ White ($\Delta$AUC = +0.06; $z = 2.34$, $p < 0.05$). Against FairCLIP-All, CMAC-MMD achieved significant improvements in five subgroups, with the largest gains observed in Female 0--60 White ($\Delta$AUC = +0.17; $p < 0.01$) and Male 0-60 White ($\Delta$AUC = +0.12; $p < 0.01$). Compared to ERM, CMAC-MMD improved AUC in seven of eight subgroups, though individual subgroup comparisons did not reach statistical significance due to smaller per-subgroup sample sizes. Critically, CMAC-MMD reduced DEOdds in all eight subgroups compared to ERM, with the largest improvements in Male 0--60 Non-White (--0.095, 55\% reduction) and Male 0--60 White (--0.094, 35\% reduction).

\begin{table}[ht]
\caption{\textbf{Clinical impact assessment: reduction in missed glaucoma diagnoses by intersectional subgroup }}\label{table5}%
\begin{tabular*}{\textwidth}{@{\extracolsep\fill}lccccc@{}}
\toprule
Subgroup & $n$ (Test) & True Positives & FN (ERM) & FN (CMAC) & FN Prevented \\
\midrule
Female 0--60 W & 108 & 37 & 35 & 34 & 1 (2.9\%) \\
Female 0--60 N-W & 109 & 43 & 35 & 33 & 2 (5.7\%) \\
Female 60+ W & 266 & 119 & 86 & 81 & 5 (5.8\%) \\
Female 60+ N-W & 131 & 65 & 44 & 40 & 4 (9.1\%) \\
Male 0--60 W & 204 & 77 & 71 & 69 & 2 (2.8\%) \\
Male 0--60 N-W & 226 & 94 & 81 & 79 & 2 (2.5\%) \\
Male 60+ W & 472 & 226 & 158 & 149 & 9 (5.7\%) \\
Male 60+ N-W & 110 & 60 & 39 & 36 & 3 (7.7\%) \\
\midrule
\textbf{Total} & \textbf{1,626} & \textbf{721} & \textbf{549} & \textbf{521} & \textbf{28 (5.1\%)} \\
\botrule
\end{tabular*}
\footnotetext{Note: W, White; N-W, Non-White; FN, false negatives; Percentages indicate relative reduction in missed diagnoses within each subgroup. CMAC refers to CMAC-MMD}
\end{table}

\textbf{Quantification of Prevented Missed Diagnoses} In the ophthalmology test set ($n = 1{,}626$), the ERM baseline produced 549 false negative diagnoses (missed glaucoma cases). CMAC-MMD reduced this to 521 false negatives, preventing 28 missed diagnoses, a 5.1\% overall reduction (Table~\ref{table5}). The clinical impact was most pronounced in non-white patient subgroups, who face higher rates of undiagnosed glaucoma in real-world settings: Female 60+ Non-White patients experienced a 9.1\% reduction in missed diagnoses (4 cases prevented), while Male 60+ Non-White patients experienced a 7.7\% reduction (3 cases prevented). Among the largest subgroup, Male 60+ White ($n = 472$), CMAC-MMD correctly identified 9 additional glaucoma cases that would have been missed by the baseline. Given that glaucoma is the leading cause of irreversible blindness and disproportionately affects racial minorities, these reductions in missed diagnoses represent clinically meaningful improvements in early detection opportunities. Embedding space visualisations for all intersectional subgroups are presented (Supplementary Fig. A6--A8), consistent with the reduced $\Delta$TPR reported in the main analysis. In addition, a summary of all embedding space evidence for margin alignment is also presented (Supplementary Appendix A.4).

\subsection{External Validation Confirms Robustness and Mechanism of Action}\label{subsec9}

\textbf{Fairness Benefits Persist Under Distribution Shift} To assess whether CMAC-MMD's fairness improvements generalise beyond the training distribution, we evaluated models trained on HAM10000 using the independent BCN20000 external validation cohort ($n \approx 12{,}000$). This out-of-distribution evaluation is critical for clinical translation, as fairness interventions that overfit to training data characteristics may fail when deployed across diverse patient populations (Table~\ref{table6}). Under distribution shift, CMAC-MMD maintained its fairness advantages with minimal impact on diagnostic performance. The overall AUC decreased marginally from 0.97 (internal) to 0.76 (external), a pattern consistent with expected domain shift effects and comparable to the ERM baseline degradation (0.94 to 0.77). The difference between methods was not statistically significant ($\Delta$AUC = -0.01; 95\% CI: -0.03 to +0.01; $p = 0.42$), confirming non-inferiority of CMAC-MMD under distribution shift.

\begin{table}[ht]
\caption{\textbf{External validation on BCN20000 confirms generalisability of fairness improvements.} CMAC-MMD maintains fairness benefits under distribution shift with minimal impact on diagnostic performance. The 35\% reduction in $\Delta$TPR demonstrates that fairness gains are not artefacts of overfitting to training data characteristics.}\label{table6}%
\begin{tabular*}{\textwidth}{@{\extracolsep\fill}lcccccc@{}}
\toprule
& \multicolumn{2}{@{}c@{}}{Performance} & \multicolumn{2}{@{}c@{}}{Fairness} & \multicolumn{2}{@{}c@{}}{Criteria} \\
\cmidrule{2-3} \cmidrule{4-5} \cmidrule{6-7}%
Method & AUC & 95\% CI & DPD & $\Delta$TPR & DF & IF-$\alpha$ \\
\midrule
ERM (Baseline) & 0.77 & [0.75--0.79] & 0.35 & 0.23 & $\times$ & $\times$ \\
\textbf{CMAC-MMD} & 0.76 & [0.74--0.78] & \textbf{0.33} & \textbf{0.15} & $\checkmark$ & $\times$ \\
\midrule
\multicolumn{3}{@{}l}{Absolute change (CMAC vs.\ ERM)} & --0.02 & --0.08 & --- & --- \\
\multicolumn{3}{@{}l}{Relative improvement} & 5.7\% & \textbf{34.8\%} & --- & --- \\
\botrule
\end{tabular*}
\end{table}

Critically, the fairness benefits of CMAC-MMD persisted under distribution shift. The $\Delta$TPR remained substantially lower for CMAC-MMD (0.15) compared to ERM (0.23), representing a 35\% relative reduction that closely mirrors the 48\% reduction observed on internal validation. The DPD similarly decreased from 0.35 to 0.33. CMAC-MMD continued to satisfy the Differential Fairness criterion ($\varepsilon = 0.5$) on external data, though neither method satisfied the stricter IF-$\alpha$ criterion under distribution shift. These findings suggest that CMAC-MMD learns a more fundamental form of equitable prediction that transfers across datasets, rather than exploiting spurious correlations present only in training data. Complete subgroup-level metrics for the external validation are provided (Supplementary Table B5--B6). Furthermore, the statistical analysis for all cohorts, including external validation, is reported (Supplementary Table B7)

\section{Discussion}\label{sec:discussion}

Our multi-cohort analysis demonstrates that intersectional diagnostic disparities in medical vision-language models can be substantially reduced without compromising, and in some cases, improving overall diagnostic performance. In both dermatology and ophthalmology domains, the proposed CMAC-MMD framework consistently narrowed gaps in $\Delta$TPR  while simultaneously enhancing global discriminative ability. These gains persisted under distribution shift during external validation, confirming that the method learns robust, generalisable features rather than overfitting to source data. Critically, our approach was one of the few methods across both domains to satisfy the Differential Fairness and Intersectional Fairness-$\alpha$ criteria, indicating that equity was achieved by improving outcomes for disadvantaged subgroups rather than degrading performance universally, the ``levelling down'' phenomenon that limits the clinical viability of many existing fairness interventions \cite{McCradden2020, Xu2022}.

The observed improvements can be attributed to a fundamental shift in the target of fairness regularisation. Standard fine-tuning systematically creates what we term a ``diagnostic certainty gap'': models become statistically less confident in their predictions for underrepresented subgroups, even when aggregate accuracy metrics appear acceptable (Fig.~\ref{fig4}). In our analysis, we observed that fine-tuning frequently causes models to learn high-confidence negative predictions for minority subgroups, effectively ignoring positive cases to maximise global loss functions. Prior fairness interventions that enforce similarity in high-dimensional embedding spaces, such as DANN or FairCLIP, often fail to translate statistical fairness in latent representations into equitable clinical outcomes because they do not directly control the decision boundary~\cite{DANN, FairCLIP, Yang2024limits}. By operating directly on the distribution of diagnostic certainty rather than abstract feature representations, CMAC-MMD ensures that the functional output most relevant to clinical decision-making, the model's diagnostic confidence, is consistent across all patient subgroups.

The statistical improvements achieved translated into clinically meaningful reductions in missed diagnoses with direct implications for patient outcomes. In dermatology, reductions in sensitivity disparities correspond to significantly fewer missed malignancies in historically underdiagnosed populations, such as young women and older men (Table~\ref{table3}). This has direct consequences for survival, given the documented disparities in melanoma outcomes driven by diagnostic delays in non-White populations.~\cite{Hu2009, Dawes2016}. Similarly, in ophthalmology, our method prevented a substantial number of missed glaucoma diagnoses, with the largest relative benefits observed in non-White subgroups(Table~\ref{table5}). Given that glaucoma is a leading cause of irreversible blindness among Black Americans, interventions that ensure equitable detection are critical for preventing vision loss in communities that already bear a disproportionate disease burden~\cite{Tielsch1991, Varma2004, Sommer1991}. Beyond diagnostic performance, the certainty gap itself carries clinical consequences: expressed model confidence influences clinician trust and subsequent actions, and systematic under-confidence for specific subgroups can produce diagnostic delays even when the final classification is technically correct~\cite{Sagona2025, Liu2025fairness}.

A critical requirement for clinical translation is that fairness benefits persist when models encounter patients from new populations, a property recent studies have shown is frequently absent, with fairness performance on internal datasets exhibiting weak or even negative correlation with fairness on external data~\cite{Yang2024limits, Drukker2023}. External validation on the BCN20000 dataset directly addresses this challenge. Under distribution shift, CMAC-MMD maintained its fairness advantages (Table~\ref{table6}). This robustness likely reflects our method's target: by enforcing consistency in decision certainty distributions rather than erasing demographic information from latent features, our method is less susceptible to the covariate and prevalence shifts that cause feature-level interventions to fail upon deployment. The generalisability is further supported by consistent performance across two distinct clinical domains—dermatology and ophthalmology, with different imaging modalities, disease characteristics, and demographic structures, suggesting that the certainty gap is a fundamental issue in medical vision-language models rather than a domain-specific artefact. From an operational perspective, CMAC-MMD offers an additional advantage: demographic attributes are required only during training to compute the fairness loss and are not accessed during inference, enabling deployment in settings where such data may be unavailable or protected by regulation while preserving patient privacy~\cite{RicciLara2022}.

Several limitations frame the scope of these findings. First, the demographic categories used—age bins, binary gender, and binarised race in the ophthalmology cohort—are imperfect social constructs that do not capture the full spectrum of human diversity and may conceal significant within-group heterogeneity~\cite{RicciLara2022, Yang2024limits}. These simplifications were pragmatic decisions driven by data availability and the requirement for sufficient subgroup sample sizes to support statistically valid metric estimation, but they limit the granularity of fairness assessment. Second, while CMAC-MMD effectively mitigates the downstream effects of bias in model predictions, it is an algorithmic intervention that cannot address upstream root causes—inequities in healthcare access, representation biases embedded in training datasets, or structural factors that determine who receives imaging in the first place. Third, although external validation demonstrated robustness under distribution shift, this study remains retrospective; prospective deployments are required to determine whether improvements in diagnostic certainty translate to changes in clinician behaviour and patient outcomes in real-world workflows. Finally, regarding the hyperparameters introduced by CMAC-MMD, we emphasise that these were empirically defined for this study; analyses presented in the Supplementary Material (Supplementary Fig. A1) demonstrate that the reported fairness improvements remain robust across parameter configurations.

In conclusion, ensuring equitable diagnostic certainty across intersectional patient subgroups is a prerequisite for the safe deployment of medical AI in diverse clinical settings. The principles underlying this approach extend beyond classification: the paradigm of aligning decision-level outputs could be adapted to other high-stakes tasks such as prognostic modelling, clinical trial matching, or treatment recommendation, where confidence in outcomes must be equitable regardless of patient demographics. We suggest that, as regulatory frameworks increasingly require demonstration of equitable performance for high-risk clinical AI methods, such as ours, that achieve fairness without compromising diagnostic accuracy will be essential for responsible clinical translation.


\bibliography{sn-bibliography}% common bib file

\section*{Supplementary Information}

This article has a supplementary material document.

\section*{Funding}

This project was funded in part by the Australian Research Council (ARC) IM240100224.

\section*{Author Contributions}
Conceptualisation: Y.Z., U.N., J.K.; Methodology: Y.Z., U.N., J.K.; Software: Y.Z.; Validation: Y.Z.; Formal Analysis: Y.Z.; Investigation: Y.Z.; Data Curation: Y.Z.; Writing -- Original Draft: Y.Z.; Writing -- Review \& Editing: Y.Z., A.D., U.N., J.K.; Visualisation: Y.Z.; Supervision: A.D., U.N., J.K.; Project Administration: J.K.

\section*{Competing Interests}

The authors have no competing interests to declare.

\section*{Ethics Approval and Consent to Participate}

This study involved secondary analysis of publicly available, de-identified datasets. The HAM10000 and BCN20000 dermatology datasets and the Harvard-FairVLMed ophthalmology dataset were collected under ethics approvals obtained by the original data custodians and are available for research use under their respective data use agreements. No additional ethics approval or informed consent was required for this analysis of anonymised data.

\section*{Data Availability}

The datasets analysed in this study are publicly available from their respective repositories. The HAM10000 dermatology dataset is available from the Harvard Dataverse (\url{https://doi.org/10.7910/DVN/DBW86T}) and the ISIC Archive (\url{https://isic-archive.com}). The BCN20000 external validation dataset is available from Figshare (\url{https://doi.org/10.6084/m9.figshare.24140028}) under a Creative Commons Attribution 4.0 International (CC-BY 4.0) license. The Harvard-FairVLMed ophthalmology dataset is available from the Harvard Ophthalmology AI Lab repository (\url{https://github.com/Harvard-Ophthalmology-AI-Lab/FairCLIP}) under a Creative Commons Attribution-NonCommercial-NoDerivatives 4.0 International (CC BY-NC-ND 4.0) license for non-commercial research purposes only.

\section*{Code Availability}

The code for implementing the CMAC-MMD framework and reproducing all experiments will be made publicly available at \url{https://github.com/YPZ404/CMAC-MMD} upon acceptance. The implementation builds upon the CLIP architecture (\url{https://github.com/openai/CLIP}). Pre-trained models used for evaluation include BioMedCLIP (\url{https://huggingface.co/microsoft/BiomedCLIP-PubMedBERT_256-vit_base_patch16_224}), PMC-CLIP (\url{https://huggingface.co/ryanyip7777/pmc_vit_l_14}), PubMedCLIP (\url{https://huggingface.co/flaviagiammarino/pubmed-clip-vit-base-patch32}), and MedCLIP (\url{https://github.com/RyanWangZf/MedCLIP}). The FairCLIP baseline implementation is available at \url{https://github.com/Harvard-Ophthalmology-AI-Lab/FairCLIP}. Results were analysed and visualised using Python v.3.12.3, NumPy v.1.24, SciPy v.1.11, and Matplotlib v.3.8.0.

\end{document}